\documentclass{aicom2e}
\usepackage{latexsym}
\usepackage{amssymb}
\newtheorem{theorem}{Theorem}[section]

\newtheorem{definition}[theorem]{Definition\rm}

\newtheorem{example}{\it Example\/}

\def \tq {\ |\ }
\def\ie{{\it i.e.\ }}
\def\etal{{\it et al.\ }}
\def \shuf {\mbox{$ \lor \hspace{-1,8 mm} \lor$}}

\def\N{\mbox{I\hspace{-.15em}N}}

\def\P{\mbox{I\hspace{-.15em}P}}

\def \ssi {\Longleftrightarrow}

\newcommand{\ici}[3]{\put(#1,#2){#3}}

\newcommand{\ENS}[3]{ $\{#1_{#2},\ldots, #1_{#3}\}$}

\newcommand{\joi}[1]%
{%
\unitlength=1mm%
\begin{picture}(3.5,2.4)%
  \put(0,0){\Large$%
\bowtie$}%
\put(0,-1.5){  \scriptsize  #1}%
\end{picture}%
}
\begin{document}
\begin{frontmatter}  
\title{Temporal Reasoning without Transitive Tables}
\runningtitle{Temporal reasoning without transitive tables (Rapport interne 2004 LIPN/LCR)}
\maketitle
\author{\fnms{Sylviane R.} \snm{Schwer}}%
\runningauthor{S. R. Schwer}
\address{LIPN UMR 7030 (Universit\'e Paris 13 et CNRS)\\sylviane.schwer@lipn.univ-paris13.fr }
\begin{abstract}
Representing and reasoning about qualitative temporal information is an
essential part of many artificial intelligence tasks.
Lots of models have been proposed in the litterature for representing such
temporal information. All derive from a point-based  or an interval-based
framework.
  One fundamental reasoning task that arises in applications of
these frameworks is given by the following scheme: given possibly
indefinite and incomplete knowledge of the binary relationships between
some temporal objects, find the consistent scenarii between all these
objects.
All these models require transitive tables --- or similarly inference rules
--- for solving such tasks.

In \cite{Sch97}, we have defined an alternative model, renamed in \cite{Sch02}  S-languages --~for Set-languages~--
 to represent qualitative temporal 
information, based on the only two relations of \emph{precedence} and \emph{simultaneity}.
In this paper, we show how this model enables to avoid
transitive tables or inference rules to handle this kind of problem.
\end{abstract}

\end{frontmatter}

\section{Introduction}
Representing and reasoning about qualitative temporal information is an
essential part of many artificial intelligence tasks. These tasks
appear in such
diverse areas as natural language processing, planning, plan
recognition, and diagnosis.
Allen \cite{All81, All83} has proposed an interval algebra framework and Vilain
and Kautz \cite{ViK86} have proposed a point algebra framework for
representing such qualitative information.
  All models that have been proposed afterwards in the
litterature derive from these two frameworks.
Placing two intervals on the Timeline, regardless of their length,
gives thirteen relations, known as
Allen's \cite{All83} relations.
Vilain \cite{Vil82} provided relations for points and intervals,
Kandrashina \cite{Kan83} provided
relations for points, intervals and chains of intervals.
Relations between two chains of intervals
 have been studied in depth by Ladkin who named them non
convex intervals \cite{Lad86}.
Al-Khatib \cite{Kha94} used a matricial approach.
Ligozat \cite{Lig91} has studied relations between chains of points,
named generalized intervals.

  One fundamental reasoning task that arises in applications in
these frameworks is given by the following scheme: given possibly
indefinite and incomplete knowledge of the relationships between
some temporal objects, find the consistent scenarii between all these
objects.
All these models have in common that the
representations of temporal
information are depicted as sets of binary relationships and are
viewed as binary constraint networks.
The reasoning is then based on transitive tables that describe the
composition of any two binary relations. 
All these models require transitive tables - or similarly inference rules
- for solving such tasks. 
The logical approach of the I.A. community explains this fact.

  The framework of formal
languages, inside which the model of S-languages has been proposed
\cite{Sch97, Sch02}, provides the same material
both for expressing the temporal objects and the n-ary relationships
between them. The reasoning is based on three
natural extensions of very well-known rational operations on
languages: the intersection, the shuffle and the projection.
More precisely, we have shown in \cite{Sch02} that binary
relations between two generalized intervals
 are in a natural correspondence with S-languages that
express Delannoy paths of order 2. By the
way, we provide to Henry Delannoy (1833-1915) a large domain of applications (though unexpected) of
his theory of minimal paths of the queen from one
corner to any other position on a chess-board \cite{Del95}.

The main idea for using formal languages for temporal representation and reasoning is that 
a word can be viewed as a line, with an arrow from left to right (the way of reading in european languages). Hence, assigning a letter to each temporal object, as its identity, and using as many occurrences of this identity as it has points or interval bounds, it is possible to describe  an atomic temporal relation between $n$ objects on the timeline, as far as there is no simultaneity, with a word on an $n$-alphabet (alphabet with $n$ letters). Simultaneity requires to be able to write several letters inside a same \emph{box}. This is the aim of the theory of S-languages. 

In this paper, we show how the S-languages framework allows to represent $n$-ary qualitative temporal relations and to reason without any transitive tables.

In the next part, we recall the basis of formal languages, following \cite{Aut87, Gin66}, and S-languages, and we examine the usual operations of the relational algebra \cite{LaM88} in the context of S-languages. 
We then provide two examples of how to reason  without transitivity tables. The first one is a revisitation of the well-known unsatisfiable closed network of Allen \cite{All83} . The second one revisits the 
Manna-Pnuelli's problem of the allocation of a resource between several requesters \cite{MaP83}. This aims to show how a problem of concurrency for complex systems, written in  modal temporal logic can be solved with the S-languages framework.

\section{Formal languages}

Let us first recall some basis on formal languages.
\subsection{Basis}
An alphabet $X$ is a finite nonempty set of symbols called letters.
 A word (of length $k \geq 0$) over an alphabet $X$ is a finite sequence $ x_1, \ldots, x_k$ of letters in $ X$.
 A word $ x_1, \ldots, x_k$ 
is usually written $ x_1 \ldots x_k$.
The unique word having no letter, \ie of length zero, called the {\it  empty} word, 
is denoted by $\varepsilon$.
The length of a word $f$ is denoted by $ |f|$. 
The number of occurrences of a letter $a$ in the word $f$ is denoted by $ |f|_a$. 
 The set of all words (resp. of length $n$) on $X$ is denoted by $X^*$ (resp. $ X^n$).
 Let us remark that  $X^* = \bigcup _{ n\geq 0} X^n$.
 The set of all words on $X$ is 
written $X^*$. A subset of $X^*$ is called a language. The empty set   $\emptyset$ is the 
least language and $X^*$ is the greatest language for the order of inclusion.

Let $u$ and $v$ be words in $X^*$.
If $u=u_1 \ldots u_r$ and $v=v_1 \ldots v_s$ are words, then $u.v$ (usually written $uv$), called
the {\it concatenation} of $u$ and $v$, is the word $u_1 \ldots u_rv_1 \ldots v_s$.
 For instance let $X=\{ x, y \}$, $u = xx$ and $v = yy$,  then the concatenation is $uv = xxyy$. Let us notice that $uv\neq vu$.
We also have to set $u^0=\{\varepsilon \}$ , $u^1=u$, $u^{n+1}=uu^n$. 
One has $v.\varepsilon = \varepsilon .v=v$. 

The concatenation can be extended to languages on $X$ by setting
$L.L'=\{ uv | u\in L, v\in L' \}$. This operation endows $2^{X^*}$ with a structure of non-commutative monoid.
We also have $L^0=\{\varepsilon \}$ , $L^1=u$, $L^{n+1}=LL^n$, $L^* = {\bigcup}_ { n\geq 0} L^n$.

$u^* = {\bigcup}_ { n\geq 0} u^n$. Even if $u^*$ is a set, it can be worked with like an element, so that we will take this alternative and use $u^*$ as a word or S-word.

 The shuffle is a very useful operator which is used in concurrency applications.
The shuffle operator  describes all possibilities of doing two concurrent
 sequences of actions in a sequential manner. Therefore, this is not a binary combination of $X^*$ because, 
from two words, it provides a set of words, that is a language. 
Its definition 
is the following: Let $u$ and $v$ be two words written on an alphabet $X^*$. 
The shuffle of $u$ and $v$ is the language $u \shuf v=
 \{ \alpha_1\beta_1  \ldots \alpha_k\beta_k \in X^* 
\vert  \alpha_1, \beta_k \in X^*, \alpha_2, \ldots , \alpha_k, \beta_1, \ldots ,  \beta_{k-1} \in X^+, u= \alpha_1 \ldots \alpha_k$, $v= \beta_1 \ldots \beta_k 
\} $.
For instance let $X=\{ x, y \}$, then  $xx \shuf yy = 
\{ xxyy, xyxy, yxxy, xyyx, yxyx,   yyxx \}$. 

 The concatenation $uv$ means an order between $u$ and $v$, this is a
word of the language $u \shuf v$.
 One has $ \varepsilon \shuf v =v \shuf \varepsilon = {v}$  
 for any word $v$ of $X^*$. The shuffle can be naturally extended to languages on $X$ by setting 
  $L \shuf L' = \bigcup_{u \in L, v\in L'} u\shuf v$.
The shuffle endows $2^{X^*}$ with a structure of commutative monoid.

Words are 
read from left to right, so that the reading induces a natural arrow of Time. 
Any 
occurrence of a letter can be viewed as an instant numbered by its position inside the word.
Traces languages or paths expressions \cite{CaH74} are used for such a purpose: planning the order of execution 
of events. But with these languages, it is not possible to differentiate two occurrences 
that are concurrent (\ie one may be {\em before}, {\em at the same time} or {\em after} the other) 
from those that must occur at the same {\em time}: these two events are said to commute.
It is presupposed that the granularity of the time measurement is 
fine enough to avoid the case {\em at the same time}.

\subsection{S-alphabet, S-words, S-languages}
In order to model explicitly concurrency with words, various tools have been proposed such as 
event structures or equivalence relations on words \ie traces.
In those theories, it is not possible to model only synchronization. 
One is able to say that two events can be done at the same time but it is not possible to express that they have
to be done at the same time. This is due to the fact that concurrency is modelled inside
a deeply sequential framework, hence, synchronization is simulated with commutativity. 
But one has to handle with \emph{instant}, in the sense of Russell \cite{Rus14}.
This is why we introduce the concept of S-alphabet which is a powerset
 of a usual alphabet.
\subsubsection{Basic definitions}

  Let us set
\begin{definition}
   If $X$ is an alphabet, an {\em  S-alphabet over $X$} is a non-empty subset of 
$2^X- \emptyset$.
 An element of an S-alphabet is an {\em S-letter}. A word on an S-alphabet is an {\em S-word}.
A set of words on an S-alphabet is an {\em S-language}.
\end{definition}
 
S-letters are written either horizontally or vertically:
$\widehat{\{ a,b \} }=\{ a, b, \left\{  {a \atop b}
\right\} \}$.  For S-letters with only two letters, we also write $ {a \choose b}$ instead of $\left\{  {a \atop b}
\right\} $ .

Examples of S-alphabets over X are:
\begin{enumerate}
\item the {\it natural} one $\dot{X} = \{ \{ a \} |\  a  \in X \} $
 that is identified with $X$.
\item the {\it full} S-alphabet over $X$, \ie $\widehat{X}= 2^X- \emptyset$.
 \item S-alphabets obtained from others S-alphabets with the following construction:\\
For an S-alphabet $Y$ over $X$, define $\overbrace{Y} =\{ A | \exists A_1, \ldots , A_k \in Y :
A = \bigcup_{i=1}^k A_i \}$.  $\overbrace{Y}$ is also an S-alphabet over $X$.
\end{enumerate}
Note that, for all S-alphabets $Y$ and $Z$  over $X$, we have
$\overbrace{\overbrace{Y}}=\overbrace{Y}$ and
$\overbrace{\widehat{Y} \cup Z}=\overbrace{Y \cup Z}$.
A S-word on a full S-alphabet over $X$ will be simply designed by an S-word on $X$. 

 In this work, we use the full S-alphabet $\widehat{X}= 2^X- \emptyset$. 
Identifying any singleton with its letter, we write
  $X \subset \widehat{X}$ and $X^* \subset {\widehat{X}}^{^*}$. 

In order to link S-words on $X$ with letters of $X$, we set
\begin{definition}
Let $X=$\ENS{x}{1}{n} be an n-alphabet and $f\in
{\widehat{X}}^{^*}$. We note
$\| f \|_{x}$ for $x\in X$ the number of occurrences of $x$ appearing inside the
S-letters of $f$, and
$\| f\|$ the integer
$\sum_{1\leq i \leq n} \| f \|_{x_i}$.
The {\em Parikh vector} of $f$, denoted $\vec{f}$, is the n-tuple
$(\| f \|_{x_1}, \ldots,\| f \|_{x_n})$.
\end{definition}
\begin{example}: \label{ExMot}

  $ f =
\left \lbrace \begin{array}{c}a \\ b \end{array}  \right \rbrace
c
b
a
\left \lbrace \begin{array}{c}a \\ c \end{array}  \right \rbrace
c
\left \lbrace \begin{array}{c}a \\  b \end{array}  \right \rbrace
a
\left \lbrace \begin{array}{c}a \\  b \\ c \end{array}  \right \rbrace
a
a
a
a
$
is an S-word such that
   $\vec{f}=(10, 4, 4)$.
\end{example}

\subsubsection{Concatenation and shuffle}
The concatenation of two S-words or two S-languages are defined exactly in the same way as in formal languages. The S-shuffle has to be generalized in the following way:

\begin{definition}
Let $X$ and $Y$ be two disjoint alphabets, $f \in
{\widehat{X}}^{^*}$, $g\in {\widehat{Y}}^{^*}$. The
{\em S-shuffle} of $f$ and $g$ is the language
$[f || g]= \{ h_1\ldots h_r\vert h_i
\in
\widehat{{X} \cup {Y}}$, with $max(|f|,|g|) \leq r \leq |f| +|g|$ and such
that there are decompositions of
$f$ and
$g$:
$f=f_1\ldots f_k$,  $g=g_1\ldots
g_k$, satisfying, 
(i) $\forall i \in [r]$, $ \vert f_i\vert, \vert g_i\vert \leq
1$,
(ii)
  $1\leq |f_i|+|g_i|$,
  and (iii)
$h_i=f_i\cup g_i 
\}.$
\end{definition}
  For instance $[aa || bb]$=$\{ aabb$,
  $a\left\{  {a \atop b}\right\} b$,
  $abab,$
  $\left\{  {a \atop b}\right\} ab$ ,$
  ab \left\{  {a \atop b}\right\}$,
  $\left\{  {a \atop b}\right\} \left\{  {a\atop
b}\right\}$,
  $ baab$,
$ba \left\{  {a \atop b} \right\} $,
   $abba$,
$\left\{  {a \atop b} \right\} ba$,
  $baba$,
$b \left\{  {a \atop b} \right\} a$,
  $bbaa \}= \{ f \in \widehat{\{ a,b \} }^* | \vec{f}=(2,2) \}$. \\

The S-shuffle of two S-languages $L$ and $L'$ written on two disjoint
alphabets is the language $[L|| L' ]=
\cup_{f\in L, f'\in L'} [f || f' ]$.

 The S-shuffle is, like the shuffle, an associative and commutative operations, which allows to note $[u_1||\cdots ||u_n]$ for the S-shuffle of $n$ S-words or S-languages.

In the case where all S-words of a language $\mathcal{L}$ share the same Parikh vector, we note 
$\vec{\mathcal{L}}$ this common Parikh vector. In particular, on the $n$-alphabet $X=\{x_1, \cdots, x_n\}$, the language,
 $$\mathcal{L}(p_1, \cdots, p_n)=\{ f \in {\hat{X}}^* |  \vec{f}=(p_1, \cdots, p_n)\}$$
  that we call $(p_1, \cdots, p_n)$-Delannoy Language -- on $X$ --, are of a particular interest for temporal qualitative reasoning, as we will show it in the next section. Let us just recall \cite{Sch02} that the cardinality $D (p_1, \ldots, p_n)$ of a $(p_1, \cdots, p_n)$-Delannoy Language is given by the following functional equation:
  $$
D (p_1, \ldots, p_n) = \sum_{\mathcal{P}red(p_1,\ldots,p_n)}
 D_n(\widetilde{p_1}, \cdots, \widetilde{p_n})
 $$
   where  for $p>0)$,  $\widetilde{p}=\{p, p-1\}$; $\widetilde{0}=\{0\}$ and
 $\mathcal{P}red((p_1, \cdots , p_n)) = 
\{ (\widetilde{p_1}, \cdots, \widetilde{p_n}) \} - \{ (p_1, \cdots , p_n)\}$. 
   
   In particular 
    $
  D(p,q)$=$ D(p,q-1) + D(p-1, q-1) + D(p-1, q) 
$
 with the initial steps
 $D(0,0)=D(0,1)=D(1,0)=1$. 
  
     Like Pascal's table for computing binomial numbers, there is a 
  Delannoy table for computing Delannoy numbers, given in Table \ref{tab2del}.
   \begin{center}
\begin{table}[h]
$\begin{array}{|c||c|c|c|c|c|c|c|c|c|}
  \hline
  {\huge _p\  ^q}&0&1&2&3&4&5&6&7&8\\
\hline\hline
0&{\bf 1}&1 &1  &1   &1   & 1& 1& 1&1\\
\hline
1&1&{\bf 3}&5  &7   &9   &11&13&15 &17\\
\hline
2&1&5 &{\bf 13}&25  &41  &61&85 &113 &145\\
\hline
3&1&7 &25 &{\bf 63}  &129 &231 & 377 & 575 & 833 \\
\hline
4&1&9 &41 &129 &{\bf 321} &681 &1289 &2241  &3649\\
\hline
5&1&11&61 &231 &681 &{\bf 1683} &3653 & 7183 & 13073\\
\hline
6&1&13&85 &377 &1289 &3653 & {\bf 8989} & 19825 & 40081\\
  \hline
7&1&15&113&575 &2241 & 7183  &  19825 & {\bf 48639} & 108545\\
\hline
8&1&17&145&833 &3649  & 13073 & 40081 & 108545 & {\bf 265729}\\
\hline
9&1&19&181&1159&5641  & 22363 & 75517 &  224143& 598417 \\
\hline
\end{array}$
\caption{\label{tab2del}Delannoy table}\end{table}
  \end{center}

     $D(p,q)$ are the  well-known Delannoy numbers \cite{Del95,Weis, Sloa} that enumerate Delannoy paths in a (p,q)-rectangular chessboard. A Delannoy path  is given as a path that can be drawn on 
a rectangular grid,
starting from the southwest corner, going to the northeast corner, 
using only three kinds of
elementary steps: {\it north}, {\it east}, and {\it north-east}.
Hence they are minimal paths with diagonal steps. The natural correspondence  between (p,q)-Delannoy paths and $\mathcal{L}(p, q)$ on the alphabet $\{a, b\}$  is: 
the S-letter $a$ corresponds to a {\it north}-step, the S-letter $b$ to a {\it east}-step and the S-letter ${ a \choose b}$ to  {\it north-east}-step.

\subsubsection{Projection}
We  extend the well-known notion of projection in formal languages theory
to S-languages. The aim is to be able to erase all occurrences of some letters in an S-word
and having as results a new S-word. The problem is how to handle an S-letter with all its
letters erased. For that purpose we set: 
 let $X$ be an alphabet and $f \in
\widehat{X}^{^*}$,
   $X_f=\{ x \in X \vert \|f\|_x \not= 0\}$. 
\begin{definition}
Let $X$ be an alphabet and $Y\subseteq X$.
The S-projection from $\widehat{X}^{^*}$ to $\widehat{Y}^{^*}$is a monoid morphism $
\pi^X_Y$ defined by the image of the S-letters:
for $ s \subset X$,
  its image is $ \pi^X_Y (s) = s
\cap Y $ if this intersection is not empty, $\varepsilon$ if not.
\end{definition}
 The projection on
$Y$ of an S-word $f$   is denoted
$ f_{|Y}$ instead of $ \pi^X_Y (f)$.

\begin{example}[Example \ref{ExMot} continued]\ \\
Let $f=\left \lbrace \begin{array}{c}a \\ b \end{array}  \right \rbrace
c
b
a
\left \lbrace \begin{array}{c}a \\ c \end{array}  \right \rbrace
c
\left \lbrace \begin{array}{c}a \\  b \end{array}  \right \rbrace
a
\left \lbrace \begin{array}{c}a \\  b \\ c \end{array}  \right \rbrace
a
a
a
a$,\\
-- $f_{|\{a\} }
=aaaaaaaaaa$,\\
-- 
$f_{|\{b\} }
$=$bbbb$,\\
--
$ f_{|\{c\} }
$=$cccc$,\\
--
$f_{|\{a, b\} }=
\left \lbrace \begin{array}{c}a \\ b \end{array}  \right \rbrace
b
a
a
\left \lbrace \begin{array}{c}a \\  b \end{array}  \right \rbrace
a
\left \lbrace \begin{array}{c}a \\  b  \end{array}  \right \rbrace
a
a
a
a$,\\
--
$f_{|\{b, c\} }= b
c
b
c
c
b
\left \lbrace \begin{array}{c}  b \\ c \end{array}  \right\rbrace 
$,\\
--
$f_{|\{a, c\} }$=
$ 
a
c
a
\left \lbrace \begin{array}{c}a \\ c \end{array}  \right \rbrace
c
a
a
\left \lbrace \begin{array}{c}a \\   c \end{array}  \right \rbrace
a
a
a
a$,\\
--
$f_{|\{a, b, c\} }$=$f$.
\end{example}
\section{Qualitative Temporal Objects and Relations in the binary algebra and their transitivity tables}
We examine qualitative temporal objects and relations inside the framework of relational algebra, as Ladkin and Maddux initiated it \cite{LaM88}. We recall the usual qualitative temporal binary algebra: the point algebra, the interval algebra \cite{All83}, the point-interval algebra \cite{Vil82, ViK86}, chains algebra. 

In this paper, we use the term {\em situation} for the description of a
unique temporal relation (complete information) between objects, which
is sometimes called an atomic relation.

Let us recall the principia of transitivity table. 
Given a particular theory $\Sigma$ supporting a set of
mutually exhaustive and pairwise disjoint dyadic situations, three individuals, $a$,
$b$ and $c$ and a pair of dyadic relations $R_1$ and $R_2$ selected from $\Sigma$
such that $R_1(a,b)$ and $R_2(b,c)$, the transitive closure $R_3(a,c)$ represents
a disjunction of all the possible dyadic situations holding between $a$ and $c$
in $\Sigma$. Each $R_3(a,c)$ result can be represented as one entry of a matrix 
for each $R_1(a,b)$ and $R_2(b,c)$ ordered pair. If there are $n$ dyadic situations 
supported by $\Sigma$, then there will be $n \times n$ entries in the matrix. 
  This matrix is a transitivity table. Transitive tables for binary
 situations have been written
for convex intervals (Allen), for points, for points and convex intervals: knowing an atomic relation between objets $A $ and $B$ and
an atomic relation between objects $B$ and $C$, derive the possible - or other said not prohibited - relations between objects $A$ and $C$.
  
Cohn \etal \cite{RCC92} have studied transitivity tables for reasoning in both time and space in a more general context. They noted the difficulty to build such secure transitivity tables.
  \subsection{The Point Algebra.}
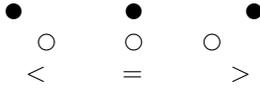
\begin{figure}[h]
	\begin{center} 
\setlength{\unitlength}{4mm}
\caption{\label{Poi} situations of the black point with respect to the white point. }
\begin{picture}(11,3)
\multiput(0.4,1)(4,0){3}{\circle*{0.5}}

\put(1.5,0){\circle{0.5}}
\put(4.4,0){\circle{0.5}}
\put(7,0){\circle{0.5}}
\put(0.8,-1.3){$<$}
\put(4,-1.3){$=$}
\put(7.6,-1.3){$>$}

\end{picture}
\end{center}
\end{figure}

The three situations are the three basic temporal relations: before $<$, equals $=$ and after $>$ as shown in Figure \ref{Poi}.  
The set of point qualitative temporal binary relations is the set
: $\{ <, >, =, \leqslant , \geqslant , \neq ,\bot ,\top\}$, where $\bot$ is the empty relation (no feasible relation) and $\top$ the universal relation (any relation is feasible). The transitive table is given in Figure \ref{fig1}
\begin{figure}[ht]
\begin{center}
$$\begin{array}{|c||c|c|c|}
\hline
\circ & < &  = & >\\
\hline
\hline
< & < & < & \top\\
\hline
=& < & = & >\\
\hline
> & \top & > & >\\
\hline
\end{array}$$
\caption{point transitivity table}
\label{fig1}
\end{center}
\end{figure}

For instance, if  $A<B$ and $B>C$ then~$A\top C$. 

\subsection{ The Interval Algebra.}
\begin{figure}[h]
\begin{center} 
\setlength{\unitlength}{0.28cm}
\begin{picture}(22,17)
\linethickness{0.15cm}
\multiput(0,0)(0,2){7}{\line(1,0){5}}
\put(10.05,15){\line(1,0){1}}
\put(11.3,14.75){{\em Rel}}
\put(14.6,14.75){OR}
\put(18,14.75){$ Rel^{\sim}$}
\put(21,15){\line(1,0){1}}
\thinlines
\put(13.5,14.75){\framebox(1,0.5)}						\put(16.7,14.75){\framebox(1,0.5)}				
\put(0.05,0.25){\framebox(4.95,0.5)}
\put(11,0.15){{\it equals}($=$) OR {\it equals}($=$)}
\put(0.05,2.25){\framebox(3,0.5)}
\put(11,2.15){{\it started-by} ($s^{\sim}$) OR {\it starts} ($s$)}
\put(1.05,4.25){\framebox(3,0.5)}
\put(11,4.15){{\it contains}($d^{\sim}$) OR {\it during} ($d$)}
\put(2,6.25){\framebox(3,0.5)}
\put(9,6.15){{\it finished-by} ($f^{\sim}$) OR {\it finishes} ($f$)}
\put(3.05,8.25){\framebox(4,0.5)}
\put(9,8.15){{\it overlaps} ($o$)OR{\it overlapped-by}($o^{\sim}$)}
\put(5.05,10.25){\framebox(3.5,0.5)}
\put(11,10.15){{\it meets} ($m$) OR {\it met-by} ($m^{\sim})$}
\put(6.05,12.25){\framebox(3.5,0.5)}
\put(11,12.15){{\it  before} ($<$) OR {\it after} ($>$)}
\end{picture}
\caption{\label{AllenRel}  the set of 13 situations between two intervals on line}
\end{center}
\end{figure}
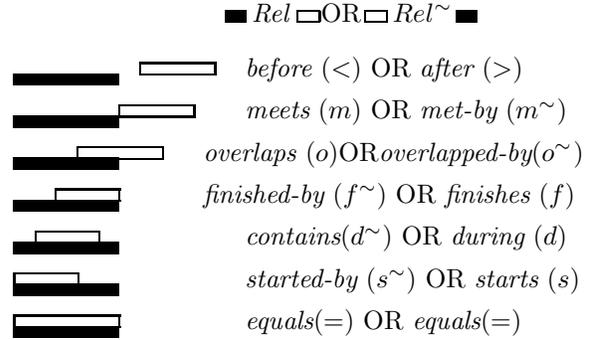

In Figure \ref{AllenRel},  we recall the thirteen situations between two intervals studied by Allen~\cite{All81}.  

The transitivity table is given in Table \ref{TAllen} where\footnote{The notation is taken, whenever possible, from Delannoy paths draw as kind of greek letters on the chess, with the following convention: Upper case for 5-subsets, lower case for 3-subsets}: \\
 $\Box=\top$, that is there is no constraint, every situation is allowed,\\
  $\Diamond=\top - \{ <, m, m^{\sim}, >\}$, which means that the two intervals intersects on an interval, this is exactly what Kamp named the overlapping\footnote{In french, couverture} relation $\circ$ on two processes \cite{Kam79} and Freksa the contemporary relation \cite{Fre91},\\
  $\Gamma^{\sim}=\{ <, m, o, s, d\}$, that is the relation \emph{begin before},\\
   $\Lambda=\{ <, m,o, d^{\sim},f^{\sim}\}$, that is the \emph{relation end after}\\
   $\alpha=\{<, m, o\}$,
    $\delta^{\sim}=\{ o, s, d\}$, 
    $\rho^{\sim}=\{o^{\sim}, d,f\}$,\\
     $\hat{s}=\{s, =, s^{\sim}\}$, 
     $\hat{f}=\{f, =, f^{\sim}\}$\\
      and using the following property: \\$\forall A \subseteq \top,  x \in A\ iff \ x^{\sim}\in A^{\sim}$.

\begin{figure}[h]
\begin{center}
$$\begin{array}{|c||c|c|c|c|c|c|c|c|c|c|c|c|c|}
 \hline
\circ & =&  < & > & d & d^{\sim} &o & o^{\sim} & m & m^{\sim} &s& s^{\sim} &f & f^{\sim} \\
\hline\hline
=& =&  < & > & d & d^{\sim} &o & o^{\sim} & m & m^{\sim} &s& s^{\sim} &f & f^{\sim} \\
\hline
< & < & < & \Box  & \Gamma^{\sim} & < & < & \Gamma^{\sim} & < & \Gamma^{\sim} & < & < & \Gamma^{\sim} & < \\
\hline
> & > & \Box & >  & \Lambda^{\sim} & > &  \Lambda^{\sim} &>  &  \Lambda^{\sim} & > &   \Lambda^{\sim}& > & > & >\\
\hline
d & d & <  &  > & d & \Box & \Gamma^{\sim} & \Lambda^{\sim} & < & > & d & \Lambda^{\sim} & d & \Gamma^{\sim}\\
\hline
d^{\sim} & d^{\sim} & \Lambda  & \Gamma  & \Diamond & d^{\sim} & \rho & \delta & \rho & \delta & \rho & d^{\sim}  & \delta & d^{\sim} \\
\hline
o & o & < & \Gamma  & \delta^{\sim} & \Lambda & \alpha & \Diamond & < & \delta & o & \rho & \delta^{\sim} & \alpha\\
\hline
o^{\sim}& o^{\sim} & \Lambda & >  & \rho^{\sim} & \Gamma & \Diamond & \alpha^{\sim} & \rho & > & \rho^{\sim} & \alpha^{\sim} & o^{\sim} & \delta\\
\hline
m & m & < & \Gamma & \delta^{\sim} & < & < & \delta^{\sim} & < & \hat{f} & m &  m & \delta^{\sim} & <\\
\hline
m^{\sim} & m^{\sim} & \Lambda & > &  \rho^{\sim} & > &\rho^{\sim} & > & \hat{s} & > & \rho^{\sim} & > & m^{\sim} &  m^{\sim} \\
\hline
s & s & < & >  & d & \Lambda & \alpha & \rho^{\sim} & < & m^{\sim} & s & \hat{s} & d & \alpha\\
\hline
s^{\sim} & s^{\sim} & \Lambda & >  & \rho^{\sim} & d^{\sim} & \rho &o^{\sim}& \rho & m^{\sim} & \hat{s} & s^{\sim} & o^{\sim} & d^{\sim} \\
\hline
f & f & < & >  & d & \Gamma & \delta^{\sim} & \alpha^{\sim} & m & > & d & \alpha^{\sim} & f & \hat{f}\\
\hline
f^{\sim} & f^{\sim} & < & \Gamma  & \delta^{\sim} & d^{\sim} & o & \delta & m^{\sim} &\delta& o &d^{\sim} &  \hat{f} & f^{\sim}\\
\hline
\end{array}$$
\caption{interval transitivity table}
\label{TAllen}
\end{center}
\end{figure}

 \subsection{The Point-Interval Algebra}
 
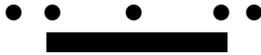
\begin{figure}[h]
\setlength{\unitlength}{4mm}
\begin{center} 

\begin{picture}(11,3)
\multiput(0.4,1)(4,0){3}{\circle*{0.5}}

\put(1.70,1){\circle*{0.5}}
\put(7.30,1){\circle*{0.5}}
\linethickness{0.25cm}
\put(1.5,0){\line(1,0){6}}

\end{picture}
  \caption{\label{IntPoi} The 5 situations between a point and an interval (each point designs a situation)}

\end{center}
\end{figure}

 In order to take into account both instantaneous and durative processes,
 Vilain provided a model with points and intervals  \cite{Vil82}.
  Figure \ref{IntPoi}  shows the  five
situations between a point and an interval.
  
 Besides the two preceding transitivity tables, are needed six more transitivity tables:\\ 
 (i) points/intervals-intervals/points, \\
 (ii) points/intervals-intervals/intervals,\\ 
 (iii) points/points-points/intervals,\\ 
 (iv) intervals/intervals-intervals/points,\\ 
 (v) inter\-vals/points-points/intervals, \\
 (vi) intervals/points-points/points.
 
\subsection{ Chains Algebras}
  
  The T-model of  Kandrashina \cite{Kan83} has three qualitative basic notions: the point, the interval and the sequence of intervals. Situations between two sequences of intervals are derived from situations between intervals. Some frequent situations are shown like the one in Figure \ref{Chain}:

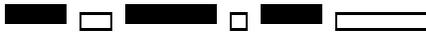
\begin{figure}[h]
\setlength{\unitlength}{4mm}
\begin{center} 

\setlength{\unitlength}{4mm}
\begin{picture}(15,2)
\linethickness{0.25cm}
\put(0.5,0.5){\line(1,0){2}}
\put(4.5,0.5){\line(1,0){3}}
\put(9,0.5){\line(1,0){2}}
\thinlines
\put(3,0){\framebox(1,0.5)}
\put(8,0){\framebox(0.5,0.5)}
\put(11.5,0){\framebox(3,0.5)}
 \end{picture}
  \caption{\label{Chain}  $S_1$ {\em alternates} $S_2$ }

\end{center}
\end{figure}
 
 These objects has been revisited and studied for their own by Ladkin \cite{Lad86} under the name of non-convex intervals.  Ligozat \cite{Lig91}  generalized to sequences of points and/or intervals under the name of   generalized intervals.  

 There are 3 situations between two points, 5 between a point and an interval, 13 situations between two intervals, 8989 situations between two sequences of three intervals or two sequences of 6 points.
  Ladkin \cite[Theorem1]{Lad86}, proved the number of situations between two chains of intervals
 is at least exponential in the number of intervals.
The exact number of situations between a sequence of $p$ points and a sequence of $q$ points has been provided by \cite[p. 83]{BeL89} without doing the connection with Delannoy numbers.
  
  Freksa studied transitivity tables with respect to convex set of intervals \cite{Fre91}, Randel \& al. \cite{RCC92} have studied transitivity tables for reasoning in both time and space in a more general context, both  in order to below the complexity rate of the computations. That was also the aim of Vilain \etal who have studied the fragment of the interval algebra, that can be written without disjunction inside the point algebra, based on the fact that  relations between intervals can be translated in terms of their bounds, inside the point algebra.  An interval $A$ is a couple of its bounds $(a,\bar{a})$ viewed as points they can contain or not, with the constraint  $a < \bar{a}$. Situations between intervals are represented in terms of the situations of their bounds:
 \begin{itemize}
\item A is {\it before} B iff $a<\bar{a}<b<\bar{b}$
 \item A {\it meets} B iff $a<\bar{a}=b<\bar{b}$
 \item A {\it overlaps} B iff $a<b<\bar{a}<\bar{b}$
 \item A {\it starts} iff B $a=b<\bar{a}<\bar{b}$
\item A {\it during} iff $b<a<\bar{a}<\bar{b}$
 \item A {\it finishes} iff  $b<a<\bar{a}=\bar{b}$
 \item A equals B iff $a=b < \bar{a}=\bar{b}$.
\end{itemize}
\section{Qualitative Temporal Objects and Relations in the S-languages  framework}
\subsection{Temporal Objects}
All temporal items previously reviewed are based on points or maximal
convex interval, that is
isolated points or pairing points. The idea is to assign an identity to each temporal objects. The set of these identities is the alphabet on which the S-languages will be written. A temporal object with identity $a$ and $p$ bounds and/or isolated points is depicted by the (S-)word $a^p$. To distinguish between points and bounds, it is possible to mark the right bound of an interval. If one non-marked letter follows a non-marked letter, then the first one depicts a point. For instance the S-word  $aa\bar{a}aaa\bar{a}a\bar{a}$ depicts the sequence : point, interval, point, point, interval, interval.

\subsection{Temporal Relations}
A relation between $n$ temporal items, using alphabet $X=\{ x_1,
\cdots, x_n\}$, is an S-word on
${\widehat{X}}^*= 2^X - \emptyset$ that describes exactly the
situation of the points on the timeline, described by
the relation. For instance
\begin{example}[Examples 1 and 2 continued]
Let A, B, C three temporal items as depicted in figure \ref{exemple}.
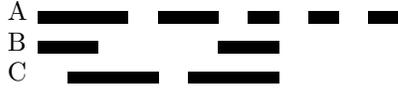
\begin{figure}
\setlength{\unitlength}{4mm}
\begin{center}
\begin{picture}(13,3)
\linethickness{0.15cm}
\put(0,-0.1){C}
\put(2,0){\line(1,0){3}}
\put(6,0){\line(1,0){3}}
\put(0,0.9){B}
\put(1,1){\line(1,0){2}}
\put(7,1){\line(1,0){2}}
\put(0,1.9){A}
\put(1,2){\line(1,0){3}}
\put(5,2){\line(1,0){2}}
\put(8,2){\line(1,0){1}}
\put(10,2){\line(1,0){1}}
\put(12,2){\line(1,0){1}}

\end{picture}
\caption{\label{exemple} A relation between 3 chains of intervals}
\end{center}
\end{figure}

On the alphabet $X = \lbrace a, b, c \rbrace$, 
item A is written aaaaaaaaaa, item B  bbbb and
item C  cccc.
The situation between them is given by the S-word  f of Example \ref{ExMot},
that is\\
$ f =
\left \lbrace \begin{array}{c}a \\ b \end{array}  \right \rbrace
c
b
a
\left \lbrace \begin{array}{c}a \\ c \end{array}  \right \rbrace
c
\left \lbrace \begin{array}{c}a \\  b \end{array}  \right \rbrace
a
\left \lbrace \begin{array}{c}a \\  b \\ c \end{array}  \right \rbrace
a
a
a
a
$.
Hence, in Example 2, we have computed:\\
-- $f_{|\{a\} }$, which is item A,\\
-- 
$f_{|\{b\} }$, which is item B,\\
--
$ f_{|\{c\} }
$, which is item C,\\
--
$f_{|\{a, b\} }$, which is the relation between A and B, \\
--
$f_{|\{b, c\} } $, which is the relation between B and C,\\
--
$f_{|\{a, c\} }$, which is the relation between A and C,\\
--
$f_{|\{a, b, c\} }$, which is the relation between A, B and C.
\end{example}

The following theorem \cite{Sch02} is the most important for our purpose:
\begin{theorem}
For any integer $n \geq 1$, let  $T_1, \cdots, T_n$ be n temporal items,
$X_n= \{ x_1 , \ldots, x_n \}$ be an  alphabet and
  $x^{p_1}, \cdots,x^{p_n}$ -- writing
$x^{n}$ the word $\underbrace{x\dots x}_{n times} $ -- their temporal
words on $X$.
  Let us denote by
  $ \Pi (p_1, \ldots, p_n)$ the set of all
  n-ary situations among $T_1, \cdots, T_n$, ${\cal L}(p_1,
\cdots , p_n)$ is its
corresponding language.
\end{theorem}

In dimension 2, it is obvious to see that there is a natural correspondence between ${\cal L}(p,q)$  and Delannoy paths in a (p,q)-rectangular chessboard.  The correspondence between interval situations, ${\cal L}(p,q)$ and (2,2)-Delannoy paths is shown in Figure \ref{Allen}, inside the N\"okel Lattice \cite{Nok90}. 

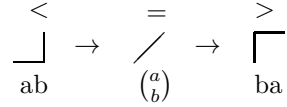
\begin{figure}[h]
\setlength{\unitlength}{4 mm}

\begin{center}

\begin{picture}(9,2.6)

\put(0,1){\line(1,0){1}}
\put(1,1){\line(0,1){1}}
\put(2,1.3){$\rightarrow$}
\put(4,1){\line(1,1){1}}
\put(6,1.3){$\rightarrow$}
\put(8,2){\line(1,0){1}}
\put(8,1){\line(0,1){1}}
\ici{0.2}{0}{ab}
\ici{0.5}{2.5}{$<$}
\ici{4.1}{0}{${a \choose b}$}
\ici{4.4}{2.4}{$=$}
\ici{8}{0}{ba}
\ici{8}{2.5}{$>$}
\end{picture}
\caption{\label{Thue1} Point lattice}

\end{center}
\end{figure}

\begin{figure*}[!htb]
\setlength{\unitlength}{3.9mm}
\begin{center}
\begin{picture}(34,20)
\put(0,10){\line(1,0){2}}
\put(2,10){\line(0,1){2}}
\put(2.5,11){\vector(1,0){1}}
\put(0,9){aabb}
\put(0.4,8){($<$)}
\put(4,10){\line(1,0){1}}
\put(6,11){\line(0,1){1}}
\put(5,10){\line(1,1){1}}
\put(6.5,11){\vector(1,0){1}}
\put(4,9){$a{a \choose b}b$}
\put(4.4,8){($m$)}
\put(8,10){\line(1,0){1}}
\put(10,11){\line(0,1){1}}
\put(9,11){\line(1,0){1}}
\put(9,10){\line(0,1){1}}
\put(8,9){abab}
\put(8.4,8){($o$)}
\put(10.5,12.5){\vector(1,1){1}}
\put(10.5,9.5){\vector(1,-1){1}}
\put(12,14){\line(1,1){1}}
\put(14,15){\line(0,1){1}}
\put(13,15){\line(1,0){1}}
\put(12,13){${a \choose b}$ab}
\put(12.4,11.8){($s$)}
\put(14.5,16.5){\vector(1,1){1}}
\put(14.5,13.5){\vector(1,-1){1}}
\put(13,7){\line(1,1){1}}
\put(13,6){\line(0,1){1}}
\put(12,6){\line(1,0){1}}
\put(12,5){ab${a \choose b}$}
\put(12.1,3.6){($f^{\sim}$)}
\put(14.5,8.5){\vector(1,1){1}}
\put(14.5,5.5){\vector(1,-1){1}}
\put(16,18){\line(0,1){1}}
\put(18,19){\line(0,1){1}}
\put(16,19){\line(1,0){2}}
\put(16,17){baab}
\put(16.4,16){($d$)}
\put(18.5,17.5){\vector(1,-1){1}}
\put(16,10){\line(1,1){1}}
\put(17,11){\line(1,1){1}}
\put(15.8,9){${a \choose b}$${a \choose b}$}
\put(16.2,7.6){($=$)}
\put(18.5,12.5){\vector(1,1){1}}
\put(18.5,9.5){\vector(1,-1){1}}
\put(16,2){\line(1,0){1}}
\put(17,2){\line(0,1){2}}
\put(17,4){\line(1,0){1}}
\put(16,1){abba}
\put(16,0){($d^{\sim}$)}
\put(18.5,4.5){\vector(1,1){1}}
\put(21,15){\line(1,1){1}}
\put(20,14){\line(0,1){1}}
\put(20,15){\line(1,0){1}}
\put(20,13){ba${a \choose b}$}
\put(20.4,11.6){($f$)}
\put(22.5,13.5){\vector(1,-1){1}}
\put(20,6){\line(1,1){1}}
\put(21,7){\line(0,1){1}}
\put(21,8){\line(1,0){1}}
\put(20,5){${a \choose b}$ba}
\put(20,3.6){($s^{\sim}$)}
\put(22.5,8.5){\vector(1,1){1}}
\put(24,11){\line(1,0){1}}
\put(24,10){\line(0,1){1}}
\put(25,12){\line(1,0){1}}
\put(25,11){\line(0,1){1}}
\put(24,9){baba}
\put(24,8){($o^{\sim}$)}
\put(26.5,11){\vector(1,0){1}}
\put(29,12){\line(1,0){1}}
\put(28,10){\line(0,1){1}}
\put(28,11){\line(1,1){1}}
\put(30.5,11){\vector(1,0){1}}
\put(28,9){b${a \choose b}$a}
\put(28,7.8){($m^{\sim}$)}
\put(32,12){\line(1,0){2}}
\put(32,10){\line(0,1){2}}
\put(32,9){bbaa}
\put(32.4,8){($>$)}
\end{picture}
\caption{\label{Allen} The N\"okel Lattice for the interval Algebra}
\end{center}
\end{figure*}
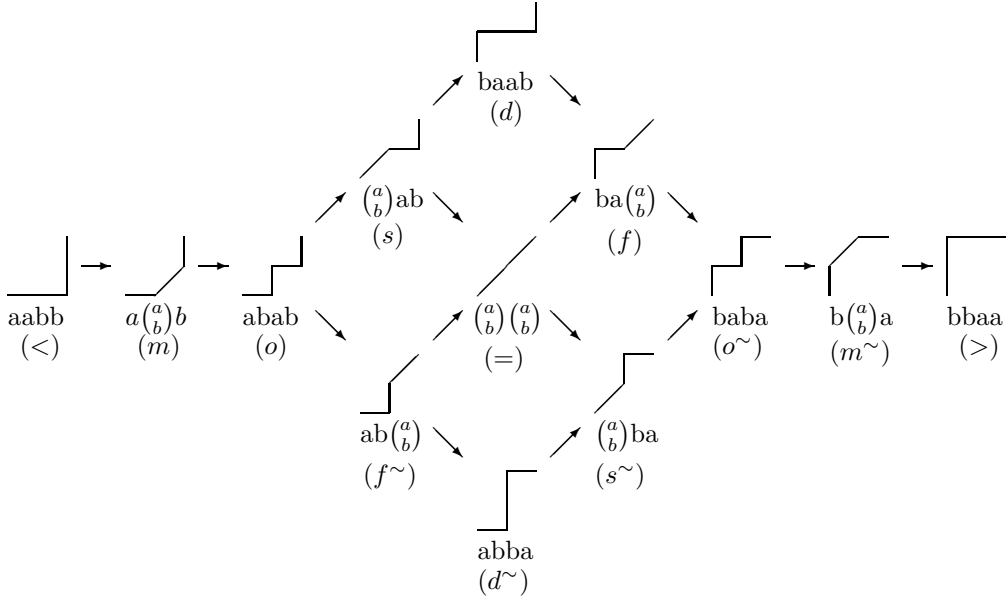

The arrow means, for S-words, the Thue rewriting rules \cite{Aut87} 
$ab \rightarrow {a \choose b }  \rightarrow ba$, which is exactly the Point lattice as we can see it in Figure  \ref{Thue1}.
Autebert \etal  have proved \cite{ALS02} that the  S-language ${\cal L}(p,q)$ on the alphabet $\{a, b\}$ (that is any set of situations between a sequence of $p$ points and a sequence of $q$ points or   Ligozat's $\Pi (p,q)$  set \cite{Lig91}) can be generated from the single S-word $a^pb^q$ and these Thue rewriting rules.  They also rigorously proved that these rules make the (p,q)Parikh vector S-language to be a distributive lattice. They also characterize the subset of $union$-irreducible S-words, which is the lattice of ideals of the 
language~:\\
$\{a^{p-1}b^{q-1}\tq  p>0,q\} || \{c\} \cup \{ a^{p-k}b^la^kb^{q-l}\tq
0<l\leq q ,0<k\leq p \} $. Its cardinality is 2$pq$.

\begin{figure}[h]
\setlength{\unitlength}{2.3mm}

\begin{center}
\begin{picture}(18,26)
\put(7.4,24){abc}
\put(7,23){\vector(-1,-1){2}}
\put(10,23){\vector(1,-1){2}}
\put(2.7,20){\{b,a\}c}
\put(3,19){\vector(-1,-1){2}}
\put(5,19){\vector(1,-2){3}}
\put(12,20){a\{c,b\}}
\put(12,19){\vector(-1,-2){3}}
\put(14,19){\vector(1,-1){2}}
\put(0,16){bac}
\put(1,15){\vector(0,-1){2}}
\put(0,12){b\{c,a\}}
\put(1,11){\vector(0,-1){2}}
\put(0,8){bca}
\put(1,7){\vector(1,-1){2}}
\put(2.3,4){\{c,b\}a}
\put(4.7,3.35){\vector(1,-1){2}}
\put(15,16){acb}
\put(16,15){\vector(0,-1){2}}
\put(14.5,12){\{c,a\}b}
\put(16,11){\vector(0,-1){2}}
\put(15,8){cab}
\put(16,7){\vector(-1,-1){2}}
\put(11.7,4){c\{b,a\}}
\put(12.5,3.35){\vector(-1,-1){2}}
\put(6.3,12){\{c,b,a\}}
\put(7.5,11){\vector(-1,-2){3}}
\put(9.5,11){\vector(1,-2){3}}
\put(7.4,0){cba}
\end{picture}

\end{center}
\caption{$\mathcal{L}(1,1,1)$ is a non modular lattice.}\label{D111}
\end{figure}
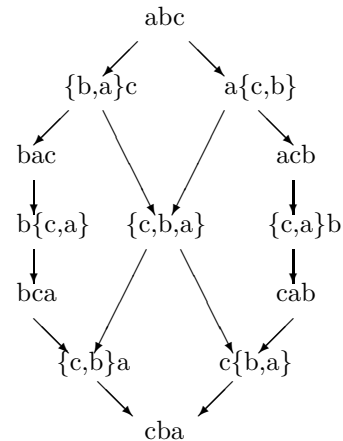

Autebert and Schwer \cite{AuS03} generalized the results to the n-ary case, proving that $\mathcal{L}(p_1, \cdots, p_n)$, with the following Thue rewriting rule is also a lattice, but not distributive because not modular, as soon as $n\geq 3$ like Figure \ref{D111} shows it. 
 Given an arbitrary order over the letters of $X$ by 
$a_1<a_2<\cdots<a_n$,
this induces over the S-letters a partial order
$P<Q\ssi[\forall x\in  P,\forall y\in Q:~x<y]$.
Then the Thue system denoted $\longrightarrow$ on
$\widehat{X}^*$, by the following:\\
$\forall P,Q,R\in \widehat{X}$ such that $P<Q$ and $R=P\cup Q$, set
$PQ\longrightarrow R$ and $R\longrightarrow QP$.
\subsection{Operations on temporal relations}
In a relational algebra, relations are basic objects on which operators operate. Apart of sets operators like union, intersection and complementation, we have yet seen two operations : the composition $\circ$ and the inverse $\sim$.  
The inverse operation exchanges the role of the objects: $a \mathcal{S} b \iff b( \mathcal{S})^{\sim} a$. There is an other unary operation, closed to the Time arrow: the
{\em symmetry} function that inverses the arrow of Time. The symmetrical  of
$a {\mathcal{S}} b$ is $a {\mathcal{S}}^{\sim} b$. 
In the framework of S-languages,
  the {\em transposition} is the identity function; the {\em symmetry} function is the mirror one that is the reading from right to left.

The third operation, the {\em composition}, is the fundamental
operation for the reasoning.
  We show now how the S-language framework avoids such a material.

These operations have their correspondents inside the S-languages framework, but we prefer to simulate them with the two new operators from S-words to S-languages that we now introduce. The first
one is the inverse of the projection, named integration, it is an unary operator; the second
one is the main operator, it is closed
to the composition of relations. It aims to answer the following question: having three worlds X, Y, Z -- with
possible intersections --, and having information $f$ in world X and information $g$ in
world Y, what possible -- i.e. not forbidden -- information can be deduced from them in world
Z? This operator is the one which allows to avoid transitive table.
\begin{definition} For any alphabet $Z$ and any word $f \in \widehat{X}^{^*}$,
\begin{itemize}
\item  The
{\em free integration} of the S-word $f$ on the alphabet $Z$, denoted $\int_Z f$, is the
S-language
$$
\int_Z f ={[\pi^{X_f\cup Z}_{X_f}]}^{-1}(f)=\{ g \in \widehat{Z\cup X_f}^{*} | g_{|X_f}=f\}
$$
\item For any distinct letters $t_1,\cdots,t_n$ of $Z$, and $\nu
=(t_1^{p_1},\cdots,t_n^{p_n}) $,  the {\em bounded to $\nu$ integration} of the S-word
$f$ on the alphabet $Z$, denoted
$\int_Z^{\nu} f$,  is the S-language
$$
\int_Z^{\nu} f=(\int_Z f )\cap (\int_{Z\cup X_f}{\cal L}(p_1, \ldots, p_n))=$$\ $$\{ g \in
\widehat{Z\cup X_f}^* | g_{|X_f}=f, \forall i: ||g||_{t_i}=p_i \}
$$
\end{itemize}

\end{definition}

These definitions are extended to languages in the following natural way:\\
The free integration of the S-language L on the alphabet $Z$, denoted $\int_Z L$,  is the
S-language $$\int_Z L = \bigcup_{f\in L} \int_Z f $$
The 
bounded to $\nu$ integration of the S-language L on the alphabet $Z$, denoted $\int_Z^{\nu}
L$,  is the S-language $$\int_Z^{\nu} L = \bigcup_{f\in L} \int_Z^{\nu} f $$

For instance - cf. end of Section 2 - let $Z=\{a,b,c\}$ and $\nu=(a^{10},b^4,c^4)$.
$\int_Z^{\nu} a^{10}=\int_Z^{\nu} b^{4}=\int_Z^{\nu} c^{4}={\cal L}_Z(10,4,4)$.

$\int_Z^{\nu} f_{|\{a, b\} }=[f_{|\{a, b\} } || c^4]$ and 
$\int_Z^{\nu} f_{|\{b, c\} }=[f_{|\{b, c\} } || a^{10}]$.

We then have

$\int_Z^{\nu} f_{|\{a, b\} } \cap \int_Z^{\nu} f_{|\{b, c\} }$=\\
$\left \lbrace \begin{array}{c}a \\  b \end{array}  \right \rbrace
cb
([aa ||cc]).
\left \lbrace \begin{array}{c}a \\  b \end{array}  \right \rbrace
a
\left \lbrace \begin{array}{c}a \\  b \\ c \end{array}  \right \rbrace
a
a
a
a$\\
 which contains the word $f$.

We never compute the integration. It is just an artifact in order to have every constraints written on the same alphabet.
The operation which costs the most is the intersection. In fact, we do not do it. We operate a kind of join, which consists in (i) computing the set of letters in common under the two integrals, (ii) verifying if all occurrences of these letters are ordered in the same manner under the two integrals (iii) shuffle the two subwords which are between two such following occurrences.  

(i) and (ii) causes no problem.
If the common letters are isolated (that is, not inside a shuffle part), the complexity of (iii) is linear but in the worse case, it can be exponential. We are studying convex part of lattices and heuristics in order to improve the complexity of the computation, in the spirit of \cite{DuS00}.

 \section{Reasoning inside the S-language framework}

It is usual in temporal applications that information arrives from many various sources or a same source can complete the
 knowledge about a same set of intervals. The usual way to deal with that, when no weight of credibility or plausibility 
is given, is to intersect all the information. The knowledge among some set of intervals interferes with some other sets of 
intervals by transitivity: if you know that Marie leaved before your arrival, and you are waiting for Ivan who attempts to
see Marie, you can tell him that he has missed her.
  
    Vilain and Kautz \cite{ViK86}  argued that there are two kinds of problems:
        \subparagraph{Problem number 1}
Let $R_1(A,C)$ and $R_2(A,C)$ be two sets of constraints between intervals A and C, what is the 
resulting set of constraints for A and C?
\subparagraph{Problem number 2}
Let $A$, $B$, $C$ be three intervals and $R(A,B)$ and $R(B,C)$ the sets of constraints
 respectively between A and B and between B and C. What is the deduced set of constraints between
A and C?

The first problem requires an {\bf and} logical operator or an {\bf intersection} set operator.
The second problem requires a transitivity operator based on tables. 

In our framework, the answers to these two problems are described exactly in the same manner, the difference being just a matter
of integration alphabet. Let ${\cal R}_1(a,b)$ [resp.${\cal R}_2(b,c)$, ${\cal R}(a,c)$ ] be the language associated to $R_1(a,b)$
[resp.
$R_2(b,c)$, ${\cal R}(a,c)$], the first answer is
$${\cal R}(a,c)= \pi^X_Z(\int_X^{(2,2,2)}{\cal R}_1(a,c)  \cap  \int_X^{(2,2,2)} {\cal R}_2(a,c))$$
and the second answer is
$${\cal R}(a,c)= \pi^X_Z(\int_X^{(2,2,2)}{\cal R}_1(a,b)  \cap  \int_X^{(2,2,2)} {\cal R}_2(b,c))$$
with in both cases $Z= \{a, c\}$.
 More generally, our main result, set in terms of intersection and integration, is:
\begin{theorem}

Let $I=\{I_1, \ldots, I_n\}$ be a set of n temporal items, $X=\{ x_1, \ldots, x_n\}$ be the corresponding alphabet and 
$\nu=(x_1^{p_1}, \cdots,x_n^{p_n})$ their Parikh vector.
Let $J_1, \ldots, J_k$ be k non empty subsets of $I$ and $Y_1, \ldots, Y_k$ their corresponding alphabets and ${\nu}_{Y_i}$ their
 Parikh vectors.
 For $ 1 \leq  i \leq  k$, let $\{ {\cal L}_{i_1}, \ldots, {\cal L}_{i_{s_i}} \} \subseteq {\cal L}({\nu}_{Y_i})$
be a set of languages describing 
$s_i$-ary temporal qualitative constraints among $J_i$.
\begin{itemize}
\item The all solution problem for I is given by the language
$$ \bigcap_{
\begin{array}{ccccc}
1 & \leq & i & \leq  & k \\
1 & \leq & j & \leq  & s_i\\
\end{array}
}
\int^{{\nu}_X}_X {\cal L}_{Ii_j}$$

\item The temporal satisfaction problem for I is satisfied if and only if 
$$ \bigcap_{
\begin{array}{ccccc}
1 & \leq & i & \leq  & k \\
1 & \leq & j & \leq  & s_i\\
\end{array}
}
\int^{{\nu}_X}_X {\cal L}_{Ii_j} \not= \emptyset$$

\end{itemize}

\end{theorem}

In order to have a look on what kind of language is computed, let us 
 revisit the unsatisfiable closed network of \cite{All83}.
\begin{figure}[h]
\setlength{\unitlength}{3mm}

\begin{center}
\begin{picture}(16,12)
\put(-0.5,0.5){A}
\put(13.5,0.5){C}
\put(6.5,3.5){B}
\put(6.5,8.5){D}
\put(6,8){\vector(-1,-1){6}}
\put(8,8){\vector(1,-1){6}}
\put(7,8){\vector(0,-1){3}}
\put(1,1){\vector(2,1){5}}
\put(13,1){\vector(-2,1){5}}
\put(1,0){\vector(1,0){12}}
\put(2,4.5){$s,m$}
\put(10.5,4.5){$s,m$}
\put(7,6.5){$o$}
\put(2.5,2.5){d,$d^{\sim}$}
\put(9.5,2.5){d,$d^{\sim}$}
\put(6,0.5){$f,f^{\sim}$ }
\end{picture}

\end{center}
\caption{Allen's instance of an inconsistent labeling.}\label{Inc}
\end{figure}
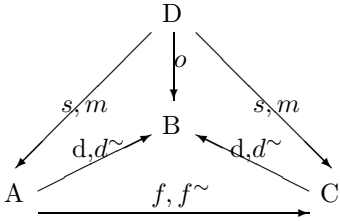
There are four intervals A, B, C, D. We then take the
 alphabet $X=\{ a, b, c, d \}$ and $\mathcal{L}(2,2,2,2)$ on $X$. 
The data are :

$$ L_1 = \int_X^{\nu}{ \lbrace \left \lbrace \begin{array}{c}a \\ d \end{array}  \right \rbrace da,
d\left \lbrace \begin{array}{c}a \\ d \end{array}  \right \rbrace a \rbrace }$$
$$L_2 = \int_X^{\nu} \{ \left \lbrace \begin{array}{c}c \\ d \end{array}  \right \rbrace dc,
d\left \lbrace \begin{array}{c}c \\ d \end{array}  \right \rbrace c \}  $$
$$L_3 = \int_X^{\nu} dbdb $$
$$L_4 = \int_X^{\nu} \{ ca  \left \lbrace \begin{array}{c}a \\ c \end{array}  \right \rbrace ,
ac \left \lbrace \begin{array}{c}a \\ c \end{array}  \right \rbrace  \}
 $$
$$L_5 = \int_X^{\nu} \{ bccb, cbbc \}$$
$$L_6 = \int_X^{\nu} \{ adda, daad  \}$$
The solution is $$L=L_1 \cap L_2 \cap L_3 \cap L_4 \cap L_5 \cap L_6$$ that we now compute.\\
$ L_1 \cap L_2 $= $$\int_X^{\nu} \{
\left \lbrace \begin{array}{c}a \\ c \\ d \end{array}    \right \rbrace 
d,
d 
\left \lbrace \begin{array}{c}a \\ c \\ d \end{array}  \right \rbrace,
\left \lbrace \begin{array}{c}a \\ d \end{array}  \right \rbrace
 \left \lbrace \begin{array}{c}c \\ d \end{array}  \right \rbrace,
\left \lbrace \begin{array}{c}c \\ d \end{array}  \right \rbrace
 \left \lbrace \begin{array}{c}a \\ d \end{array}  \right \rbrace 
\}
\lbrack a || c \rbrack
$$ or equivalently, due to the lack of occurrences of $b$, \\
$ L_1 \cap L_2 $=\\ $ \biggl[  \{
\left \lbrace \begin{array}{c}a \\ c \\ d \end{array}    \right \rbrace 
d,
d 
\left \lbrace \begin{array}{c}a \\ c \\ d \end{array}  \right \rbrace,
\left \lbrace \begin{array}{c}a \\ d \end{array}  \right \rbrace
 \left \lbrace \begin{array}{c}c \\ d \end{array}  \right \rbrace,
\left \lbrace \begin{array}{c}c \\ d \end{array}  \right \rbrace
 \left \lbrace \begin{array}{c}a \\ d \end{array}  \right \rbrace 
\}.$\\
$\lbrack a|| c \rbrack
 || bb  \biggl] $\\
This language contains 164 words.\\
$ L_1 \cap L_2 \cap L_3 $= \\$\{
\left \lbrace \begin{array}{c}a \\ c \\ d \end{array}  \right \rbrace bd,
db \left \lbrace \begin{array}{c}a \\ c \\ d \end{array}  \right \rbrace,
\left \lbrace \begin{array}{c}a \\ d \end{array}  \right \rbrace b \left \lbrace \begin{array}{c}c \\ d \end{array}  \right \rbrace,
\left \lbrace \begin{array}{c}c \\ d \end{array}  \right \rbrace b \left \lbrace \begin{array}{c}a \\ d \end{array}  \right \rbrace
\}.$\\
$
\lbrack a || b || c \rbrack $ \\
This language contains 52 words.\\
$  L_1 \cap L_2 \cap L_3 \cap L_4 $=$$
									\{
\left \lbrace \begin{array}{c}a \\ d \end{array}  \right \rbrace
b
\left \lbrace \begin{array}{c}c \\ d \end{array}  \right \rbrace ,
\left \lbrace \begin{array}{c}c \\ d \end{array}  \right \rbrace
b
\left \lbrace \begin{array}{c}a \\ d \end{array}  \right \rbrace																\}.
\biggl[ 
				\left \lbrace \begin{array}{c}a \\ c \end{array}  \right \rbrace ||
					b
\biggr]  $$ This language contains 6 words.\\
$  L_1 \cap L_2 \cap L_3 \cap L_4 \cap L_5$= $$\{
\left \lbrace \begin{array}{c}a \\ d \end{array}  \right \rbrace
b
\left \lbrace \begin{array}{c}c \\ d \end{array}  \right \rbrace  
\left \lbrace \begin{array}{c}a \\ c \end{array}  \right \rbrace
b,
\left \lbrace \begin{array}{c}c \\ d \end{array}  \right \rbrace
b
\left \lbrace \begin{array}{c}a \\ d \end{array}  \right \rbrace  
b
\left \lbrace \begin{array}{c}a \\ c \end{array}  \right \rbrace
\}$$ This language has 2 words.
$$ L_1 \cap L_2 \cap L_3 \cap L_4 \cap L_5 \cap L_6 = \emptyset$$
This language is empty: the problem is unsatisfiable.
\section{Temporal reasoning about concurrent systems}
Most of the temporal properties of programs have been studied inside either the framework 
of temporal logics or modal logic. Temporal reasoning about concurrent systems can be partitioned in a natural way into two classes related to the modality used : necessity, symbolized by $\Box$ or possibility, symbolized by $\Diamond$. Translated in a 
temporal framework, we use the terms \emph{always} and \emph{sometime}. These modalities were 
studied first by the Megarians\footnote{A school founded by Euclid of Megaric, student of Socrates, like Plato, this school was concurrent of Aristotle's one.}, then by Aristotle and the Stoic. Despite their variant, these modalities are linked to the universal $\forall$ and particular $\exists$ quantifiers.

Manna and Pnuelli \cite{MaP81, MaP83}, defined three
important classes of temporal properties of concurrent programs that are
investigated inside the modal temporal
framework. They are invariance, liveness and precedence properties. The first two ones are closely related to the two basis modalities $\Box$, which is interpreted as \emph{always  from now on} and $\Diamond$ which is interpreted as $sometimes$ or \emph{at least one time from now on}. The third one is an extension of the $\Diamond$ class, proposed by Manna and Pnuelli, closed to the $\mathcal{U}$, the \emph{until} modality in order to capture precedence in general. These three classes are available for
temporal properties in general. We first review these three classes
and then we revisit their
  example inside our formal languages framework. Our purpose is not to criticize their solution, but only to translate it inside our model.
\subsection{Temporal properties types}

The first class is the class of invariance properties. These are properties that can be expressed by a temporal formula of the form: $\Box \psi$ or $\varphi \Rightarrow \Box\psi$. Such a formula, stated for a program $P$, says that every computation of $P$ continuously satisfies $\psi$ throughout the rest of the computation either from the beginning (first formula) or  whenever $\varphi$ becomes true. Among properties falling into this class are: partial correctness, error-free behavior, mutual exclusion and absence of deadlocks.

The second set, associated to the $sometimes$ modality  defines the \emph{liveness}  properties class.
These properties are expressible by temporal formulas of the form: $\Diamond \psi$ or $\varphi \Rightarrow \Diamond\psi$. In both cases these formulas guarantee the occurrence of some event $\psi$; in the first case unconditionally and  in the second case conditional on an earlier occurrence of the event $\varphi$. Among properties falling into this class are:  total correctness, termination, accessibility,  lack of individual starvation, and responsiveness.

The third class is the precedence properties
class which is very well-known inside the artificial intelligence community. In a broad sense, Manna and Pnuelli asserted that precedence properties are all the properties that are expressible using the \emph{until} operator  $\mathcal{U}$ in formulas such as $\chi \mathcal{U} \psi$ or $\varphi  \Rightarrow \chi \mathcal{U} \psi$. In both cases the formulas again guarantee the occurrence of the event $\psi$, but they also ensure that from now until that occurrence, $\chi$ will continuously hold. Among properties falling into the \emph{until} class are strict (FIFO) responsiveness, and bounded overtaking.
The meaning they give to the precedence operator, that is to the formula $p<q$, is that $q$ eventually happens, and $p<q$ is automatically satisfied if $q$ never happens. Hence we have:
$$p<q \equiv \neg ( ( \neg p) \mathcal{U} q)$$

We are now ready to embark inside  the problem of allocating a single resource between several requesters as explained in \cite{MaP83}. In this paper we recall almost every thing about their specification, because this kind of problem is of general interest for all complex systems.
\subsection{The Allocation Problem}
  Let us consider a program $G$ (granter) serving
as an allocator of a single resource between several processes
(requesters) $ R_1, \ldots ,R_k$
competing for the resource. Let each $R_i$ communicate with $G$ by
means of two boolean variables:
$r_i$ and $g_i$. The variable $r_i$ is set to $ true$ ($=1$) by the requester
to signal a request for the
resource. Once $R_i$ has the resource it signals its release by
setting it to $false$ ($=0$).
  The allocator
G signals $R_i$ that
the resource is granted to him by setting $g_i$ to $true$. Having obtained
  a release signal from $R_i$, which is indicated by $r_i=false$, some
time later, it will appropriate
the resource by setting $g_i$ to $false$. 
\subsubsection{Properties of the system described inside the modal logic framework.}
Several obvious and important properties of this system belong to the invariance and liveness classes. 
\paragraph{An invariant property.}
Insuring that the resource is granted to at most one requester at a time is an invariant property:
$$\Box ( \sum_{i=1}^{k}g_i) \leq 1$$
\paragraph{A liveness property: responsiveness}
The important property that ensures responsiveness, i.e.  which guarantees that every request $r_i$ will
eventually be granted by setting
$g_i$ to $true$ is a liveness property:
$$  ( \forall i )  (1 \leq i \leq k) (r_i \Rightarrow \Diamond g_i )$$
\paragraph{ Precedence properties.}
Two precedence properties are set.
\subparagraph{-- An absolute precedence property: Absence of Unsolicited Response.}
An important but often overlooked desired feature is that the resource 
 will not be granted to a party who has not requested it. A similar property in the context of a communication network is that every message received must have been sent by somebody. This is expressible by the temporal formula:
$$ \neg g_i \Rightarrow (r_i < g_i)$$ 
The formula states that if presently $g_i$ is $false$, i.e., $R_i$ does not presently have the resource, then before the resource will be granted to $R_i$ the next time, $R_i$ must signal a request by setting $r_i$ to $true$.
\subparagraph{-- A relative precedence property: a Strict (FIFO) Responsiveness.}
Sometimes the weak commitment of eventually responding to request is not sufficient. At the other extreme we may insist that responses are ordered in a sequence parallelling the order of arrival of the corresponding requests. Thus if requester $R_i$ succeeded in placing in request before requester $R_j$, the grant to $R_i$ should precede the grant to $R_j$. A straightforward translation of this sentence yields the following intuitive but slightly imprecise expression: $(r_i < r_j) \Rightarrow (g_i<g_j)$. A more precise expression is
$ ( \forall i,j ) (i \ne j) (1 \leq i,j \leq k)$
$$((r_i \land \neg r_j \land \neg g_j ) \Rightarrow (\neg g_j {\cal{U}} g_i).$$
It states that if ever we find ourselves in a situation where $r_i$ is presently on,   $r_j$ and $g_j$ are both off, then we are guaranteed to eventually get a $g_i$, and until that moment, no grant will be made to $R_j$. Note that $r_i \wedge \neg r_j$ implies that $R_i$'s request precedes $R_j$'s request, which has not been materialized yet. 

We implicitly rely here on the assumption that once a request has been made, it is not withdrawn until the request has been honored.
 This assumption can also be made explicit as part of the specification, using another precedence expression:
 $$r_i \Rightarrow g_i < (\neg r_i).$$
 Note that while all the earlier properties are requirements from the granter, and should be viewed as the \emph{post-condition} part of the specification, this requirement is the responsibility of the requesters. It can be viewed as part of the \emph{pre-condition} of the specification.

Two assumptions are also implicitly used but not mentioned:
\begin{enumerate}
\item a process is not allowed to make an other
request until he has given the resource back,
\item there are as many
requests from $R_i$ as grants for $R_i$.
\end{enumerate}

We now are leaving the way Manna and Pnuelli have resolved the problem inside the modal logic framework: finding an abstract computation model based on sequences of transitions and states, a proof system \cite{MaP83}.
\subsection{revisitation of the allocation problem inside the S-languages framework}
\subsubsection{Objects and relations representations}

Our formalization attempts to translate any information into a
temporal information. The two
boolean variables introduced in the formulation of the problem do not
belong to the problem but to
one of its data interpretation. These boolean variables are evolving through the timeline. It is natural to represent the values of boolean  variables evolving through the time 
as characteristics function of boolean variables on a linear order that can be called temporal boolean functions  \cite{BoL95, JuS97, JuS02}. Inside a determined and bounded period of time, the temporal boolean functions $r_i$
(resp. $g_i$) can be interpreted
  as a chain of intervals with $n_i$
 intervals, where $n_i$ is the number of requests
(resp. grants). This number is exactly known at the end of the fixed period of time. Any interval is a maximal period inside which the value of $r_i$ (resp. $g_i$) is $true$. 

A priori, for the general specification, either we can choose to write $n_i$ as an indeterminate number or to set $*$ for saying that there is a finite but unknown number. To each requesters, we provide  $2k$ chains of intervals written on their identities  alphabet $X=\{ r_1,  g_1,  \ldots , r_k, g_k,  \}$. Any language $L$ that
satisfies the problem is such that its Parikh number is
$\vec{L} \subseteq \underbrace{ ((2,2)\N,\ldots,(2,2)\N) }_{\rm k \ times}$. But for the sake of an easier reading, we will use the following alphabet
$X=\{ r_1, \bar{r}_1, g_1, \bar{g}_1 \ldots , r_k, \bar{r}_k,g_k, \bar{g}_k \}$, which allows to make a distinction between the beginning of the interval (no marked letter) and the end of the interval (marked letter) so that  any language $L$ that
satisfies the problem is such that its Parikh number is
$\vec{L} \subseteq \underbrace{ ((1,1,1,1)\N,\ldots,(1,1,1,1)\N) }_{\rm k \ times}$.
\subsubsection{The specification.}
\paragraph{The invariant property.}
Insuring that the resource is granted to at most one requester at a time can be interpreted like this:
this constraint only concerns granters. After having read a letter $g_i$ -- a beginning of an allocation to $R_i$ --, the first following occurrence of a letter of type $g$ or $\bar{g}$ inside the S-word is necessarily a $\bar{g}_i$ letter (the end of the current allocation to $R_i$. This constraint is formulated by the S-language$$L_1= \int_X (g_{1}\bar{g_1}, \ldots , g_{k}\bar{g_k} )^*.$$  Hence  we have
$ L \subseteq L_1 $.
\paragraph{The liveness property.}
The  guaranty that  every request $r_i$ will
eventually be granted concerns each $R_i$ individually. This organizes every couple of S-words $(r_i\bar{r}_i)^*$ and $(g_i\bar{g}_i)^*$ saying that after each occurrence of a letter $r_i$, the first occurrence of a letter among the set $\{ r_i, \bar{r}_i, g_i, \bar{g}_i\}$ is an occurrence of the letter $g_i$. There is no constraint between the occurrences of the letters $\bar{r}_i$ and $\bar{g}_i$ that is to say that we get
$$
 L \subseteq  \bigcap_{1 \leq i \leq k} \int_X (r_ig_i \lbrack \bar{r_i} ,\bar{g_i }\rbrack )^*
$$
But we can give a more precise S-language, because we know that the sequence of actions
related to any request is the following:
 request, allocation, release and deallocation. So more precisely we can give 
$$ 
L \subseteq L_2 =\bigcap_{1 \leq i \leq k} \int_X (r_ig_i\bar{r_i}\bar{g_i })^*.
$$
\paragraph{ The two precedence properties.}
\subparagraph{-- Absence of Unsolicited Response.}
A resource will be not granted to a party who has not requested it.
This constraint is already written in the preceding constraint.
\subparagraph{-- Strict (FIFO) Responsiveness.}
Grants are ordered in the same order as the corresponding requests. This concerns how the occurrences of $r_i$, $r_j$, $g_i$, $g_j$ are shuffled together. The case where $r_i=r_j$ has been already taken into account in $L_2$. Restricted to the alphabet $\{ r_i, r_j, g_i, g_j\}$, if $R_i$ requests before $R_j$,  either $R_i$ is granted before the request of $R_j$ or after\footnote{The system is not allowed to receive simultaneous requests.}. But $R_j$ can request before $R_i$. That is we get four cases that can be organized as product but not as a shuffle. Hence we have
$$  L \subseteq L_3 = \bigcap_{1 \leq i <  j \leq k}
\int_X (r_ig_i, r_jg_j, r_ir_jg_ig_j,r_jr_ig_jg_i )^*$$

\subsubsection{The solution.}
All constraints are now to be specified in terms of S-languages. Every S-word
 contained in $ L_1 \cap L_2 \cap L_3 $ satisfies the problem, hence the solution is
$L =
\int_X (g_{1}\bar{g_1}, \ldots , g_{k}\bar{g_k} )^*
\cap$\ $ \bigcap_{1 \leq i \leq k} \int_X (r_ig_i\bar{r_i} \bar{g_i })^*
\cap \bigcap_{1 \leq i \ne j \leq k}
\int_X (r_ig_i, r_jg_j, r_ir_jg_ig_j,r_jr_ig_jg_i )^*$

If the system can't realize two tasks simultaneously, the solution will be
$ 
L \cap X^*
$.
\subsubsection{Example}
 Let us suppose that there are three requesters and the order of the requests is $R_1R_2R_3R_1R_3$. The temporal objects are the four chains expressed with the S-words - that are also words - $r_1\bar{r}_1r_1\bar{r}_1$,  $g_1\bar{g_1}g_1\bar{g_1}$,
$r_3\bar{r_3}r_3\bar{r_3}$, $g_3\bar{g_3}g_3\bar{g_3}$, and the intervals expressed with the S-words $r_2\bar{r_2}$, $g_2\bar{g_2}$.\ The set of all possible situations is given by the language $[r_1\bar{r_1}r_1\bar{r_1}||g_1\bar{g_1}g_1\bar{g_1}||r_3\bar{r_3}r_3\bar{r_3}||g_3\bar{g_3}g_3\bar{g_3}||r_2\bar{r_2}||g_2\bar{g_2}]$. That is the S-language with Parikh vector $((2,2)$, $(2,2)$,$(1,1)$,$(1,1),(2,2),(2,2))$ on the ordered alphabet $\{ r_1, \bar{r}_1,g_1,  \bar{g}_1$, $r_2, \bar{r}_2,g_2,  \bar{g}_2,r_3, \bar{r}_3,g_3,  \bar{g}_3\}$.

The requests order $R_1R_2R_3R_1R_3$ induces the following sequence between the letters of type $r$ : $r_1r_2r_3r_1r_3$.\\
 $L_2=\int_X \{ r_1g_1\bar{r_1}\bar{g_1}r_1g_1\bar{r_1}\bar{g_1}$, $r_2g_2\bar{r_2}\bar{g_2}$, $r_3g_3\bar{r_3}\bar{g_3}\-r_3g_3\bar{r_3}\bar{g_3}\}$.\\
 The condition on the $r$ letters and the expression of s$L_3$ allows to substitute to $L_3$, the language $L_3'$:\\ 
 $L_3'=\int_X g_1g_2g_3g_1g_3$.\\
 $L_1 =\int_X g_1\bar{g_1}g_2\bar{g_2}g_3\bar{g_3}g_1\bar{g_1}g_3\bar{g_3}$.\

 The shuffle of all these fragments are depicted by a Hasse graph of the precedence ordering on the instants corresponding to the bounds of intervals, that is the letters in figure \ref{Manna}. This is a graphical representation of the resulting S-language $L$.

\begin{figure*}[htb]
\setlength{\unitlength}{8mm}
\caption{\label{Manna} Graphs of the example resulting S-language}
\begin{center}
\begin{picture}(16,3.3)
\put(0.5,3){$r_1$}
\put(3.5,3){$r_2$}
\put(6.5,3){$r_3$}
\put(9.5,3){$r_1$}
\put(12.5,3){$r_3$}
\put(1,3.1){\vector(1,0){2.3}}
\put(4,3.1){\vector(1,0){2.3}}
\put(7,3.1){\vector(1,0){2.3}}
\put(10,3.1){\vector(1,0){2.3}}
\put(1.9,0){$\bar{r_1}$}
\put(4.9,0){$\bar{r_2}$}
\put(7.9,0){$\bar{r_3}$}
\put(10.9,0){$\bar{r_1}$}
\put(13.9,0){$\bar{r_3}$}
\put(1,1.5){$g_1$}
\put(4,1.5){$g_2$}
\put(7,1.5){$g_3$}
\put(10,1.5){$g_1$}
\put(13,1.5){$g_3$}
\put(2.7,1.5){$\bar{g_1}$}
\put(5.7,1.5){$\bar{g_2}$}
\put(8.7,1.5){$\bar{g_3}$}
\put(11.7,1.5){$\bar{g_1}$}
\put(14.7,1.5){$\bar{g_3}$}
\put(3.2,1.6){\vector(1,0){0.7}}
\put(6.2,1.6){\vector(1,0){0.7}}
\put(9.2,1.6){\vector(1,0){0.7}}
\put(12.2,1.6){\vector(1,0){0.7}}
\put(0.7,2.8){\vector(1,-2){0.5}}
\put(3.7,2.8){\vector(1,-2){0.5}}
\put(6.7,2.8){\vector(1,-2){0.5}}
\put(9.7,2.8){\vector(1,-2){0.5}}
\put(12.7,2.8){\vector(1,-2){0.5}}
\put(1.4,1.3){\vector(1,-2){0.5}}
\put(4.4,1.3){\vector(1,-2){0.5}}
\put(7.4,1.3){\vector(1,-2){0.5}}
\put(10.4,1.3){\vector(1,-2){0.5}}
\put(13.4,1.3){\vector(1,-2){0.5}}
\put(2.3,0.3){\vector(1,2){0.5}}
\put(5.3,0.3){\vector(1,2){0.5}}
\put(8.3,0.3){\vector(1,2){0.5}}
\put(11.3,0.3){\vector(1,2){0.5}}
\put(14.3,0.3){\vector(1,2){0.5}}
\put(3.1,1.7){\line(5,1){6.2}}
\put(9.2,2.9){\vector(4,1){0.2}}
\put(9.1,1.7){\vector(3,1){3.5}}

\end{picture}
\end{center}
\end{figure*}
\section{Conclusion}
In this paper, we have presented the S-languages framework and shown how to represent and to reason on qualitative temporal problem.
The Hasse Diagram we provide for the allocation problem has to be compared with the temporally labeled graph of Gerevini and Schubert \cite{GeS95}. 
 
Different implementations have been made in order to improve the complexity on the computations. They take benefits of the algorithms used for computing operations in formal languages theory, and the use of automata theory. The problems come, as usual, from the parts of S-languages that have to be broken into disjoint parts, in order to go on in the computation. 

Two implementations has already been made, concerning the interval algebra. The first one is based on the notion of pattern \cite{DuS00}. The second one \cite{Pic03}  applies it on the linguistical model of Descl\'es based  on topological intervals \cite{Des90}. Bounds of intervals are labelled in order to mention whether an interval is open or closed.
As $ \mathcal{L}(p_1, \cdots, p_n)$ is a lattice, we work on convex parts, following the approach of Freksa \cite{Fre91}. 
These prototypes suggest that ii may be better to compute in two steps: first accepting all situations, even those not allowed situations,  and second without forbidden situations, rather than to  compute directly the good solution. 

\section*{Acknowledgements}
I would thank Jean-Michel Autebert and J\'er\^ome Cardot for their fruitful comments on the text.

\end{document}